# Markov Chain Monte Carlo using Tree-Based Priors on Model Structure

Nicos Angelopoulos    James Cussens
Department of Computer Science, University of York
Heslington, York, YO10 5DD, UK
{nicos,jc}@cs.york.ac.uk

## Abstract

We present a general framework for defining priors on model structure and sampling from the posterior using the Metropolis-Hastings algorithm. The key ideas are that structure priors are defined via a probability tree and that the proposal distribution for the Metropolis-Hastings algorithm is defined using the prior, thereby defining a cheaply computable acceptance probability. We have applied this approach to Bayesian net structure learning using a number of priors and proposal distributions. Our results show that these must be chosen appropriately for this approach to be successful.

## 1 INTRODUCTION

This paper extends and empirically evaluates a general framework for Bayesian modelling which was briefly sketched in (Cussens, 2000). Our primary goal is to implement a *practical* method of incorporating prior information about model structure. The current lack of such a method is noted by Friedman and Koller (2000) in their paper on Bayesian learning of Bayesian network structure which is closely related to this paper

> ... relatively little attention has been paid to the choice of structure prior, and a simple prior is often chosen largely for pragmatic reasons.... The standard priors over network structure are often used not because they are particularly appropriate for a task, but rather because they are simple and easy to work with.

Although our approach is very general—for example, in (Cussens, 2000) it was applied to a model space composed of logic programs—the experiments here are focused exclusively on learning Bayesian network (BN) structure from data and prior knowledge. In many applications we are likely to have at least some knowledge about network structure which we are willing to model as hard constraints, for example, that $X$ is/is not a parent/child/ancestor/descendant of $Y$, that $X$ and $Y$ are independent/dependent, that no family has more than $k$ parents, etc. We may also wish to express softer prior beliefs, for example, encoding a preference for sparsely connected networks. Both hard and soft prior information are expressible in our method.

## 2 TREE-BASED PRIORS ON MODEL STRUCTURE

Our approach to defining priors on a finite or countably infinite space of models is best understood in terms of the sampling process which selects a model from the model space. The sampling process is a series of *independent* choices where for each choice-point there is a multinomial distribution over the choices available. Some sequences of choices are defined as *successful* and these determine (or *yield*) a model in the space. Sequences of choices can be represented as distinct branches in a *probability tree*. Figure 5 (see last page) presents $BNTREE$, an example of such a probability tree for the very small model space of all 25 Bayesian networks consisting of the random variables $B$, $L$ and $S$. At each choice point in this tree we choose how to connect a particular pair of variables where the pairs are considered in the following order: $(B, L), (L, S), (B, S)$.

$BNTREE$ is very simple in that each choice-point has the same multinomial distribution. If $(X, Y)$ is the pair of random variables under consideration then there is probability $p_1$ that $Y$ is chosen to be a parent of $X$, probability $p_2$ that $Y$ is chosen to be a child of $X$, and probability $p_3$ that there is to be no direct connection between $X$ and $Y$. In general, we could have many multinomials, with the $p_i$ depending on $(X, Y)$, thus giving a very detailed prior specification.

Two branches, or *derivations* as we shall now call them, in $BNTREE$ lead to cyclic graphs at leaves 2 and 13— which are not in the model space of Bayesian nets. These are therefore labelled as *failure derivations* and correspond to failed attempts to sample a Bayesian net. If a derivation is not a failure derivation then it is a *successful derivation*.

$BNTREE$ defines three probability distributions: $\psi_\lambda$, $f_\lambda$ and $p_\lambda$. $\lambda$ denotes the log versions of the parameters, so for $BNTREE$, $\lambda = (\log p_1, \log p_2, \log p_3)$. For any derivation $x$, $\psi_\lambda(x)$ is the product of the probabilities attached to the choices that constitute $x$. For example if $x_2$ is the derivation leading to leaf 2 in $BNTREE$, then $\psi_\lambda(x_2) = p_1^2 p_2$. $f_\lambda(x)$ is simply $\psi_\lambda(x|x \text{ is successful})$, so:

$$f_\lambda(x) = \begin{cases} Z_\lambda^{-1} \psi_\lambda(x) & \text{if } x \text{ successful} \\ 0 & \text{otherwise} \end{cases}$$

where $Z_\lambda$ is the normalising constant:

$$Z_\lambda = \sum_{x \text{ is successful}} \psi_\lambda(x)$$

From a sampling point of view, $Z_\lambda$ is the probability that the next attempt at sampling will be successful. $f_\lambda$ is essentially a log-linear distribution (also know as a Gibbs distribution, or, in the computational linguistics literature, a MAXENT distribution). To see this let the frequency of a choice $C_i$ in a successful derivation $x$ be $\nu_i(x)$, then $f_\lambda(x) = Z_\lambda^{-1} \prod_i p_i^{\nu_i(x)} = Z_\lambda^{-1} \exp(\sum_i \lambda_i \nu_i(x))$. In the language of log-linear modelling, the $\nu_i$ are the 'features' of derivations. Here the features are integer-valued but this is not generally the case for log-linear models.

In the particular case of $BNTREE$ there is only one derivation for each BN, so there is a bijection between models and successful derivations. In general, this is not the case. To see this, consider the probability tree $CGTREE$ in Figure 1 which defines a distribution $f_\lambda$ over 4 successful derivations and a distribution, which we will call $p_\lambda$, over the model space which consists of 3 chain graphs. At each choice point we choose between either an edge from $A$ to $B$ or an edge from $B$ to $A$. An undirected edge corresponds to an edge in both directions. Two derivations, $x_2$ and $x_3$, yield the undirected graph $M_{A-B}$. We define $p_\lambda$ so that $p_\lambda(M_{A-B}) = f_\lambda(x_2) + f_\lambda(x_3) = p_1 p_2 + p_2 p_1 = 2 p_1 p_2$. In general, $p_\lambda(M)$ is just the distribution over models defined by marginalising derivations away:

$$p_\lambda(M) = \sum_{x \text{ yields } M} f_\lambda(x) \qquad (1)$$

To get a feel for the distributions defined by $BNTREE$, suppose we set $p_1 = p_2 = p_3 = 1/3$. This means $\forall x \psi_\lambda(x) = 1/27$. $Z_\lambda = 25/27$, so $f_\lambda(x) = 1/25$ for all successful $x$ and $p_\lambda(M) = 1/25$ for all BNs. Increasing $p_3$ at the expense of $p_1$ and $p_2$ expresses a preference for sparsely connected BNs and would make $BN_{27}$ (the BN at leaf 27 in Fig 5) the mode of the prior. Setting $p_1 = 0$ almost surely removes each leftmost choice from each choice-point effectively reducing the model space to $(BN_{14}, BN_{15}, BN_{17}, BN_{18}, BN_{23}, BN_{24}, BN_{26}, BN_{27})$, the BNs consistent with the variable ordering $(B, L, S)$. Setting $p_3 = 0$ effectively reduces the model space to $(BN_1, BN_4, BN_5, BN_{10}, BN_{11}, BN_{14})$, the 6 totally connected BNs.

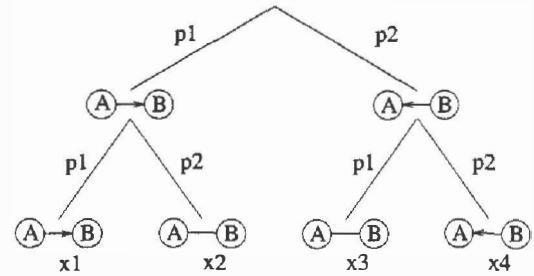

Figure 1: $CGTREE$ probability tree

## 3　MCMC USING THE METROPOLIS-HASTINGS ALGORITHM

In the Bayesian approach to statistical inference we do not 'learn' or 'induce' a single best model from the data $D$. Instead a posterior distribution $P(M|D)$ is derived from the prior distribution $P(M)$ using Bayes theorem. This posterior distribution is then used to compute the expected values of various quantities of interest.

Unfortunately, we know of no way of updating the probability trees representing our priors into (practical) trees representing posteriors. Worse still, we have no way of even sampling directly from the posterior. This is a common problem in complex Bayesian modelling and can be addressed by employing Markov chain Monte Carlo (MCMC) methods (Gilks, Richardson, & Spiegelhalter, 1996). Given a *target distribution* (e.g. a posterior) from which we would ideally like to sample but cannot, an MCMC algorithm constructs a Markov chain whose *stationary distribution* is the target distribution. If we run the chain for long enough then estimates of features of the target distribution produced by sampling this chain will converge to their true values, under fairly weak conditions.

We use the Metropolis-Hastings (MH) algorithm to construct a Markov chain $M^0, M^1, M^2, \ldots$. When the target density is a posterior $P(M|D) = P(M)P(D|M)/P(D)$, MH is defined as follows (where the initial model $M^0$ is sampled from the prior).

1. Generate a candidate value $M^*$ with *proposal* probability distribution $q(M^i, M^*)$.

2. Set $M^{i+1} = M^*$ with probability

$$\alpha(M^i, M^*) = \min\left\{\frac{q(M^*, M^i)}{q(M^i, M^*)}\frac{P(D|M^*)P(M^*)}{P(D|M^i)P(M^i)}, 1\right\} \quad (2)$$

else set $M^{i+1} = M^i$

As long as the Markov chain is *ergodic* (Gilks et al., 1996), the proposal distribution $q$ can have *any* form and the stationary distribution will be $P(M|D)$.

## 4 CHOOSING A PROPOSAL DISTRIBUTION FOR TREE-BASED PRIORS

Our proposal mechanism uses the same probability tree that we used to define the prior. We use MH to construct a Markov chain of *derivations* with stationary distribution $f_\lambda(x|D)$. This gets us to our target posterior distribution $p_\lambda(M|D)$ by marginalisation similarly to (1). In our MCMC samples, we do not even record which derivation we are visiting, just the model associated with it. (Failure derivations never appear in the Markov chain.) Given this close connection between $f_\lambda(x|D)$ and $p_\lambda(M|D)$ we will switch freely between talking about sequences of derivations and sequences of models.

The basic idea is to 'bounce' around the tree by backtracking from the current leaf to an interior node and then going down the tree to a new leaf by choosing branches using the probabilities which define the prior distribution. Note that there is a one-one correspondence between leaves and derivations, so we can describe the algorithm in terms of visiting leaves. The only caveat is that when we stop backtracking we may not choose the branch up which we have just backtracked—such branches are temporarily *blocked*. This ensures that there is a unique path between any two leaves. Using the prior distribution to construct the proposal distribution is also done in (Philps & Smith, 1996; Chipman, George, & McCulloch, 1998; Denison, Mallick, & Smith, 1998) and has the advantage that many of the terms in (2) cancel out. In particular the $Z_\lambda$ normalising constant, which is particularly difficult to compute for large trees, disappears.

Our proposal distribution is parameterised by a *backtrack probability* $p_b$ which controls the sizes of jumps. A candidate leaf $M^*$ is generated from a current leaf $M^i$ as follows.

1. Backtrack one step to the most recent choice point in the probability tree.

2. We then probabilistically backtrack as follows: If at the top of the tree go to step 3. Otherwise with probability $p_b$ backtrack one more step to the next choice point and repeat step 2, or with probability $1 - p_b$ go to step 3.

3. Once we have stopped backtracking choose a new leaf $M^*$ from the choice point by selecting branches according to their probabilities attached to them. However, in the first step down the tree we may not choose the branch that leads back to $M^i$.

To illustrate this, consider some leaf pairs in $BNTREE$ (the numbers refer to nodes in $BNTREE$):

$$\begin{aligned}
q(BN_1, BN_2) &= (1-p_b).p_2.(1-p_1)^{-1} \text{ (via 28)}\\
q(BN_1, BN_3) &= (1-p_b).p_3.(1-p_1)^{-1} \text{ (via 28)}\\
q(BN_1, BN_8) &= p_b(1-p_b).p_3 p_2.(1-p_1)^{-1}\\
&\quad \text{(via 28,37,30)}\\
q(BN_1, BN_{10}) &= p_b^2.p_2 p_1 p_1.(1-p_1)^{-1}\\
&\quad \text{(via 28,37,40,38,31)}\\
q(BN_{10}, BN_1) &= p_b^2.p_1 p_1 p_1.(1-p_2)^{-1}\\
&\quad \text{(via 31,38,40,37,28)}
\end{aligned}$$

Each proposal probability has three factors (separated by '.' above): a backtrack factor, a factor from the prior and a factor for the temporarily blocked branch.

To define the acceptance probability $\alpha(M^i, M^*)$ we need some more notation. Let $M^i$ have depth $n^i$ and $M^*$ depth $n^*$. Let $Node(M^i, M^*)$ be the deepest common ancestor of $M^i$ and $M^*$, this is the internal node we reach when we stop backtracking from $M^i$. Let $C^i$ be the choice from $Node(M^i, M^*)$ that leads to $M^i$, and define $C^*$ analogously. Let $p^i$ and $p^*$ be the probabilities attached to $C^i$ and $C^*$, respectively. Then, as proven in (Cussens, 2000), if $M^*$ is a leaf at the end of a *successful* derivation:

$$\alpha(M^i, M^*) = \min\left\{p_b^{(n^*-n^i)}\frac{1-p^i}{1-p^*}\frac{P(D|M^*)}{P(D|M^i)}, 1\right\} \quad (3)$$

If $M^*$ is a failure (for example, $BN_2$ or $BN_{13}$ in $BNTREE$) then $\alpha(M^i, M^*) = 0$.

It is easy to see that our Markov chain is ergodic, i.e. positive recurrent and aperiodic. (See (Roberts, 1996) for the basic definitions and theorems regarding ergodicity.) For any leaves $M_i$ and $M_j$, it is clear that the probability $P_{ij}(t)$ of moving from $M_i$ to $M_j$ in $t$ steps, is positive, for *all* $t > 0$. So *a fortiori* the Markov chain is irreducible and, since we are doing MCMC, also positive recurrent. The chain is also obviously aperiodic and so we have ergodicity.

## 5 BAYES FACTORS FOR BAYESIAN NETS

The data only affects MH via the quantity $P(D|M^*)/P(D|M^i)$ the ratio of marginal likelihoods for the proposed and current model. This ratio is known as a Bayes factor. In all but deterministic non-probabilistic models the full likelihood requires not just a determination of model structure but also of the model parameters. In the case of a BN network structure $M$, these are the conditional probability distributions $\theta_{X_i|\mathbf{u}_i}$ for the random variables $X_i$ and each instantiation $\mathbf{u}_i$ of their parents $\mathrm{Pa}_M(X_i)$. To calculate the Bayes factor we need to marginalise away (i.e. integrate away) these model parameters.

Here we make the standard assumptions to allow this integration to have a closed form. Firstly, we assume a Dirichlet prior over $\theta_{X_i|\mathbf{u}_i}$ for all $X_i$ and $\mathbf{u}_i$. We also assume *global parameter independence* and *parameter modularity*. The former says that the total density over the complete parameter set is the product of the individual $\theta_{X_i|\mathbf{u}_i}$ densities and the latter that $\theta_{X_i|\mathbf{u}_i}$ is the same for any two network structures where $X_i$ has the same parents. Our final assumption is that the data is complete, each data point contains an observation for each random variable.

Given all these convenient assumptions we have from (Heckerman, 1996) that the marginal likelihood is a product of 'scores' for each family in the BN $M$:

$$P(D|M) = \prod_i \mathrm{score}(X_i, \mathrm{Pa}_M(X_i)|D) \qquad (4)$$

Let $N_{ijk}$ be the count in the data for $X_i$ taking value $k$ when its parents have instantiation $j$. Let $\alpha_{ijk}$ be the corresponding Dirichlet parameter. Let $\mathrm{Dom}(X_i)$ be set of possible values of $X_i$. Define $N_{ij} = \sum_{k \in \mathrm{Dom}(X_i)} N_{ijk}$ and $\alpha_{ij} = \sum_{k \in \mathrm{Dom}(X_i)} \alpha_{ijk}$. Then $\mathrm{score}(X_i, \mathrm{Pa}_M(X_i)|D)$ is

$$\prod_{j \in \mathbf{u}_i} \frac{\Gamma(\alpha_{ij})}{\Gamma(\alpha_{ij} + N_{ij})} \prod_{k \in \mathrm{Dom}(X_i)} \frac{\Gamma(\alpha_{ijk} + N_{ijk})}{\Gamma(\alpha_{ijk})} \qquad (5)$$

If $X_i$ has the same parents in $M^i$ and $M^*$ then the score for $X_i$ cancels out in the Bayes factor making it quick to compute the Bayes factor for similar BNs. We also cache scores. All our computations are done by taking logs and using the log $\Gamma$ function from the C maths library. Everything else is done in Prolog including the definition of the $\alpha_{ijk}$. Setting the $\alpha_{ijk}$ to informative values expresses prior knowledge about likely conditional probability values, but in our current artificial experiments we set $\alpha_{ijk} = 1, \forall i, j, k$.

## 6 DEFINING TREE-BASED PRIORS USING STOCHASTIC LOGIC PROGRAMS

Our tree-based MCMC approach is implemented using stochastic logic programs (SLPs). This is because the logic

```
bn([],[],[]).
bn([RV|RVs],BN,AncBN) :-
  bn(RVs,BN2,AncBN2),
  connect_no_cycles(RV,BN2,AncBN2,BN,AncBN).

%RV parent of H, p_1=1/3
1/3 :: which_edge([H|T],RV,[H-RV|Rest]) :-
  choose_edges(T,RV,Rest).
%RV child of H, p_2=1/3
1/3 :: which_edge([H|T],RV,[RV-H|Rest]) :-
  choose_edges(T,RV,Rest).
%no edge, p_3=1/3
1/3 :: which_edge([_H|T],RV,Rest) :-
  choose_edges(T,RV,Rest).
```

Figure 2: The predicates bn/3 and which_edge/3, part of $\mathcal{S}_{BN}$, an SLP defining priors over BN structure

programming paradigm gives us a number of facilities for free:

1. Easy declarative representation of the model space and constraints on models using a subset of first-order logic;

2. Implicit definition of the tree structure (by a logic program and goal);

3. Automatic construction of models as we move down the tree (building first-order terms via successive unifications;

4. Backtracking concepts which could be extended to incorporate sampling over SLPs;

5. A built-in notion of failure (unification failure).

In most of our experiments we used variants of the prior defined by the SLP $\mathcal{S}_{BN}$ in Fig 2. The predicate bn/3 in Fig 2 says that BN is a Bayesian net constructed from the variables in the list [RV|RVs] if BN2 is a Bayesian net constructed from the list RVs and RV is connected to BN2 without introducing any cycles. The Prolog operator ':-' should be read as '←'. BNs are represented as ordered lists of families, for example $BN_8$ is represented as [b-[l],l-[],s-[b]]. This is a natural representation and one that allows us to use utility predicates from the Sicstus Prolog ugraphs library. The third argument in bn/3 represents the ancestor sets for each node, so for $BN_8$ it would be [b-[l],l-[],s-[b,l]]. The ancestor sets allow quick checking for cycles and allow us to easily put constraints on ancestor relationships.

A logic program $\mathcal{P}$ together with a *goal* $G$, defines an *SLD-tree* each branch of which is a refutation of $G$ using $\mathcal{P}$. For example, given the SLP $\mathcal{S}_{BN}$ parts of which are given in Fig 2 and the goal $\forall BN \neg bn([s,l,b], BN)$ ("there are no $BN$s with $s, l$ and $b$ as nodes") we (essentially) get $BNTREE$ as an SLD-tree. Each successful branch refutes our goal and provides an instantiation of $BN$ as a counterexample. In other words, each branch is a constructive proof that there exists a particular $BN$.

It is important to understand that the SLD-tree is never actually constructed, it is just a representation of Prolog's search space (standard Prolog explores it deterministically: depth-first, leftmost-first). This means we can define very large, even infinite trees, with a logic program.

To define a prior over models using an SLP, one takes a logic program $\mathcal{P}$ and associates probabilities with some of the clauses in the logic program. For example, in $\mathcal{S}_{BN}$ the three clauses defining `which_edge/3` are the only probabilistic ones. Denote the resulting SLP by $\mathcal{S}$ and let the (logs of) the probabilities added be $\lambda = (\lambda_1, \lambda_2, \ldots, \lambda_n)$. For any goal $G$, $\mathcal{S}$ has an associated SLD-tree: the one for its underlying logic program $\mathcal{P}$. Each choice in the SLD-tree is associated with a choice of clause, so some choices will have associated probabilities. Construct a probability tree by deleting all choices from the SLD-tree which have no associated probability. The SLP defines three distributions $\psi_{(\lambda,\mathcal{S},G)}$, $f_{(\lambda,\mathcal{S},G)}$ and $p_{(\lambda,\mathcal{S},G)}$ (usually abbreviated to $\psi_\lambda$, $f_\lambda$ and $p_\lambda$) via this probability tree exactly as described in Section 2. For a more detailed account of how SLPs define distributions see (Cussens, 2001).

## 7 EXPERIMENTS

### 7.1 A PRELIMINARY EXPERIMENT

For our first experiment we sampled 10,000 data points from the 3-node BN constructed from the $S$, $L$ and $B$ nodes of the well-known 'Asia' network given in Fig 3. This is $BN_{19}$ in Fig 5. We then used a uniform prior over the set of 25 BNs with three nodes and did MCMC as described above with $10^6$ iterations and a burn-in of 5000. The results are in Table 1. The BNs in Table 1 fall into two Markov equivalence classes divided by a horizontal line. Note that the BN that generated the data ($BN_{19}$) has much lower posterior probability than $BN_{22}$. Although the posterior probabilities estimated by MCMC are eventually close to the true values, convergence is very slow.

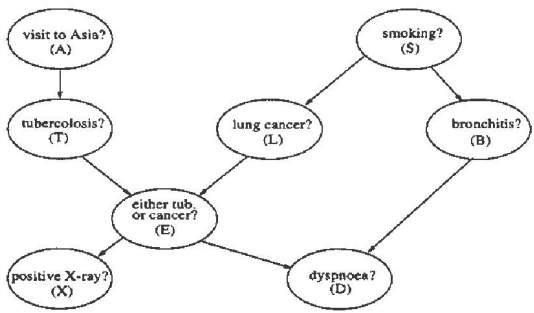

Figure 3: 'Asia' Bayesian net

| $M$ | $\hat{p}_4$ | $\hat{p}_5$ | $\hat{p}_6$ | $p$ |
|---|---|---|---|---|
| $BN_{22}$ | 0.668 | 0.690 | 0.704 | 0.702 |
| $BN_{20}$ | 0.176 | 0.150 | 0.145 | 0.146 |
| $BN_{19}$ | 0.144 | 0.152 | 0.143 | 0.145 |
| $BN_4$ | 0.007 | 0.005 | 0.005 | 0.005 |
| $BN_5$ | 0.002 | 0.001 | 0.002 | 0.002 |
| $BN_1$ | 0.001 | 0.001 | 0.001 | 0.001 |
| $BN_{14}$ | 0 | 0 | 0 | 0 |
| $BN_{10}$ | 0.001 | 0 | 0 | 0 |
| $BN_{11}$ | 0 | 0 | 0 | 0 |

Table 1: Estimated ($\hat{p}_i$) and actual ($p$) posterior probabilities for the nine most probable 3-node BNs in $BNTREE$ rounded to 3 d.p. $\hat{p}_i$ is the estimated probability after $10^i$ iterations. The other sixteen highly improbable BNs were never visited, so for them $\forall i : \hat{p}_i = 0$.

### 7.2 IMPROVING MCMC

To address the issue of slow convergence and hence produce reasonable estimates in bigger model spaces we made a number of improvements over the basic approach described above. One of the key issues in MCMC is to strike a balance between local jumps to 'neighbouring' candidate models and big jumps to distant ones. With a bias towards local jumps there is a risk that the chain will remain stuck in a particular neighbourhood, perhaps failing to visit areas of high posterior probability. This is less of a problem if we can move a big distance via a sequence of local jumps. Unfortunately, this is not the case with the trees, such as $BNTREE$, associated with priors defined by $\mathcal{S}_{BN}$ in Fig 2. Since all the leaves are of equal depth (this depth is 28 for the 8-node BN model space) we can only reach a leaf on the 'other side' of the root node by doing one big jump via this root node—there are no shallower leaves that can help us reach the root node by a sequence of small jumps. Even with $p_b = 0.98$ our chains get stuck, very rarely proposing distant candidates and rejecting them when they are proposed.

Solving this problem by going for big jumps brings its own problems. Sooner or later the chain should find its way to a model with a posterior probability higher than the vast majority of other models, including the vast majority of distant models. This will mean that many candidate models will be rejected—the chain will be stuck—unable to find other models of high posterior probability from amongst the mass of low probability models. Indeed, the rationale for local jumps is that we should give ourselves the opportunity to visit neighbouring models, since if a model has high probability then maybe so will many of its neighbours, since they are, in some way, similar.

In our approach the neighbourhoods are defined by the topology of the tree, and the distance between them is regulated by $p_b$, the backtrack probability. We effect a compromise between local and big jumps by implementing a *cyclic transition kernel*. We cycle through the values $p_b = 1 - 2^{-n}$, for $n = 1, \ldots, 28$, so that on every 28th iteration, there is a high probability of backtracking all the way to the top of the tree.

We have also found it effective where possible to manually re-write our SLPs so that they are failure-free. If there are lots of failure points in the probability/SLD tree, then many candidates will be failures. Since we never jump to a failure leaf this can lead to poor mixing, particularly when we include many constraints so that successful branches are surrounded by many failures.

### 7.3 RESULTS

We generated 2295 data points from the entire 'Asia' BN and constructed Markov chains over the space of all 8-node BNs. There are 783,702,329,343 BNs in this model space. Firstly we used an SLP which defined a prior over all 8-node BNs consistent with a particular variable ordering (one consistent with the 'Asia' BN). We also did experiments with the added constraints that the number of parents for a node is limited to $k = 2$ and $k = 3$. The first-order framework makes it easy to add constraints. However care is needed to write SLP priors which allow reasonable convergence.

For each setting of the constraints, we did two runs each of 500,000 iterations. We then used the sample so created to estimate the posterior probability that $X_i$ is a parent of $X_j$ for all $i, j$ consistent with the ordering. There are 28 of these features. Plotting the estimates from the two runs against each other produced the results in Fig 6. In Fig 6 we also have results when only half of the data was used (1124 data points). We then ran the same set of experiments but did not impose an ordering on the variables, and in one case removed the limit on the number of parents. The results of these experiments are in Fig 7. For each pair of runs, Figs 6 and 7 contain runtimes in seconds for a Pentium III 1GHz running SICStus 3.8.5 under Linux.

Ideally, all points in Figs 6 and 7 would be on the diagonal, showing that estimates produced from the two different runs were equal. Basically, we get points near the diagonal if either the estimates are near 0 or 1 (there are lots of these and they are not easily visible on the plots) or the model space is constrained, particularly if constrained by an ordering. In the totally unrestricted case (rightmost plot of Fig 7), we get a few really bad points. In the worst case $A$ was almost always the parent of $D$ in the first run, but this was almost never the case in the second. These results fit our expectations, the "near 0 or 1" phenomenon follows from basic sampling theory, and it is no surprise that estimates are more reliable in constrained settings. One surprise is that we expected better results using only half the data, since then the shape of the posterior is less spiky, however the results are essentially the same.

To see whether we could use the child-parent probability estimates to get a partial picture of the 'true' BN we constructed a BN where there was a link from $X_i$ to $X_j$ whenever that link had estimated probability greater than 95%. We did this for (i) the totally unconstrained case, (ii) when only a maximum of $k = 3$ parents were allowed and (iii) when only a maximum of $k = 3$ parents were allowed, but the parents of $E$ were forced to be $T$ and $L$. This last constraint seems reasonable since $E$ is defined to be "either $T$ or $L$". The resulting BNs are Fig 8. As expected, we get spurious links in the unconstrained case with $A$ and $T$ parents of $B$, even though $A$ and $T$ are marginally independent of $B$ in the data generating 'Asia' network. For the two cases with $k = 3$ constraints on the parents, we get no links that are not in 'Asia'. It is surprising that we never have either of $T$ or $L$ the parent of $E$ even though $E$ is a function of $T$ and $L$.

## 8  DISCUSSION

Friedman and Koller (2000) state that "different runs of MCMC over networks lead to very different estimates in the posterior probabilities of structural features". Although a number of our experiments have this problem it is not always the case (although, unlike Friedman and Koller (2000), we have yet to work with variable sets as large as 37). The consistency of the probability estimates depends on how constrained the model space is (as well as how well designed our MCMC strategy is). The stronger our prior knowledge the better the estimates. We see our main contribution as providing a framework for incorporating prior knowledge in a declarative and practical manner. One advantage of working directly on network structure is that it makes it easier to actually write down the sort of prior knowledge we might have.

It is instructive to compare our approach with the Markov chain Monte Carlo model composition ($MC^3$) approach introduced by Madigan and York (1995). The $MC^3$ algorithm moves through the model space by altering only one edge at a time. The proposal distribution is not based upon the prior, and so the ratio of the priors of the current and proposed models is computed to find the acceptance probability, in contrast to our (3). Priors are used where this ratio is easily found and $MC^3$ has been successfully applied to a number of datasets.

Madigan, Andersson, Perlman, and Volinsky (1996) point out that there are strong computational and statistical arguments in favour of using a model space each element of which is a *Markov-equivalence class of BNs*, rather than a single BN. Surprisingly, each class has on average only about 4 members—a significant proportion have only one member (Gillespie & Perlman, 2001). Each Markov-equivalence class $[D]$ is uniquely represented by a unique chain graph $D^*$, the *essential graph*. It is easy to define a prior over essential graphs and hence over Markov-equivalence classes using an SLP. These can be extended to define a prior over BNs if desired. $S_{EG}$, a fragment of which is given in Fig 4, does this and is most easily understood by looking at the steps used to sample from the prior it defines. Given a set of random variables RVs, skeleton/2 first probabilistically chooses an undirected graph Skel, then essential_graph/3 probabilistically chooses an essential graph EG with *immoralities* Imms by adding arrows to Skel. bn/3 is then defined to probabilistically choose a particular BN from the equivalence class of BNs defined by EG. It is more natural to define a prior over essential graphs, but we extended the prior to be over BNs to compare with our previous experiments.

```
bn(RVs,BN) :-
        skeleton(RVs,Skel),
        essential_graph(Skel,Imms,EG),
        bn(EG,Imms,BN),
        top_sort(BN,_).   %check for cycles
```

Figure 4: Fragment of $S_{EG}$, an SLP prior based on Markov-equivalence classes

Although it was easy to define the prior, the associated probability tree had far too many failure derivations to be useful. A similar problem occurred when, in a separate experiment, we used a prior with constraints enforcing marginal independence between pairs of nodes.

Although all the experiments here are for BNs our approach is general. Our SLP implementation, which works by translating a human-readable SLP as in Fig 2 to Prolog, does not 'know' that the first-order terms are BNs. If we want to apply our method to different models, we just write the appropriate SLP priors and write code to compute the appropriate Bayes factors.

Our final conclusion is that further work is required to fully understand the advantages and limitations of our approach. There are many methods in the literature for improving MCMC and we have only tried one of them (the use of a cyclic transition kernel). One big problem was that natural ways of defining priors generally led to inefficient MCMC due to the existence of many failures. The same problem arises in logic programming where it is addressed using source-to-source program transformation. We expect program transformation to be necessary for real applications, where, if we can exploit prior information using our framework, substantial benefits are possible.

### Acknowledgements

This work was supported by EPSRC research grant GR/N09152 *Induction of Stochastic Logic Programs*. Many thanks to our four anonymous reviewers for their insightful criticisms of a previous version of this paper.

## References

Chipman, H. A., George, E. I., & McCulloch, R. E. (1998). Bayesian CART model search. *Journal of the American Statistical Association*, *39*(443), 935–960. With discussion.

Cussens, J. (2000). Stochastic logic programs: Sampling, inference and applications. In *Proceedings of the Sixteenth Annual Conference on Uncertainty in Artificial Intelligence (UAI-2000)*, pp. 115–122 San Francisco, CA. Morgan Kaufmann.

Cussens, J. (2001). Parameter estimation in stochastic logic programs. *Machine Learning*, *44*(3), 245–271.

Denison, D. G. T., Mallick, B. K., & Smith, A. F. M. (1998). A Bayesian CART algorithm. *Biometrika*, *85*(2), 363–377.

Friedman, N., & Koller, D. (2000). Being Bayesian about network structure. In *Uncertainty in Artificial Intelligence: Proceedings of the Sixteenth Conference (UAI-2000)*, pp. 201–210 San Francisco, CA. Morgan Kaufmann Publishers.

Gilks, W. R., Richardson, S., & Spiegelhalter, D. (Eds.). (1996). *Markov Chain Monte Carlo in Practice*. Chapman & Hall, London.

Gillespie, S. B., & Perlman, M. D. (2001). Enumerating Markov equivalence classes of acyclic digraph models. In *Uncertainty in Artificial Intelligence: Proceedings of the Seventeenth Conference (UAI-2001)* San Francisco, CA. Morgan Kaufmann Publishers.

Heckerman, D. (1996). A tutorial on learning with Bayesian networks. Tech. rep. MSR-TR-95-06, Microsoft Research.

Madigan, D., Andersson, S. A., Perlman, M. D., & Volinsky, C. T. (1996). Bayesian model averaging and model selection for Markov equivalence classes of acycic digraphs. *Communications in Statistics: Theory and Methods*, *25*, 2493–2520.

Madigan, D., & York, J. (1995). Bayesian graphical models for discrete data. *International Statistical Review*, *63*, 215–232.

Philps, D. B., & Smith, A. F. M. (1996). Bayesian model comparison via jump diffusions. In Gilks, W. R., Richardson, S., & Spiegelhalter, D. (Eds.), *Markov Chain Monte Carlo in Practice*, pp. 215–240. Chapman & Hall, London.

Roberts, G. O. (1996). Markov chain concepts related to sampling algorithms. In Gilks, W. R., Richardson, S., & Spiegelhalter, D. (Eds.), *Markov Chain Monte Carlo in Practice*, pp. 45–57. Chapman & Hall, London.

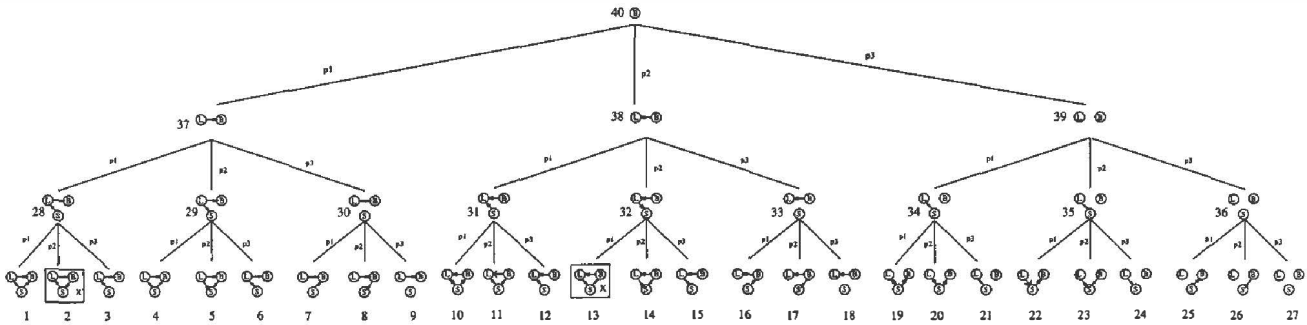

Figure 5: $BNTREE$ probability tree defining a distribution over all Bayes nets consisting of the three random variables $A, B, C$.

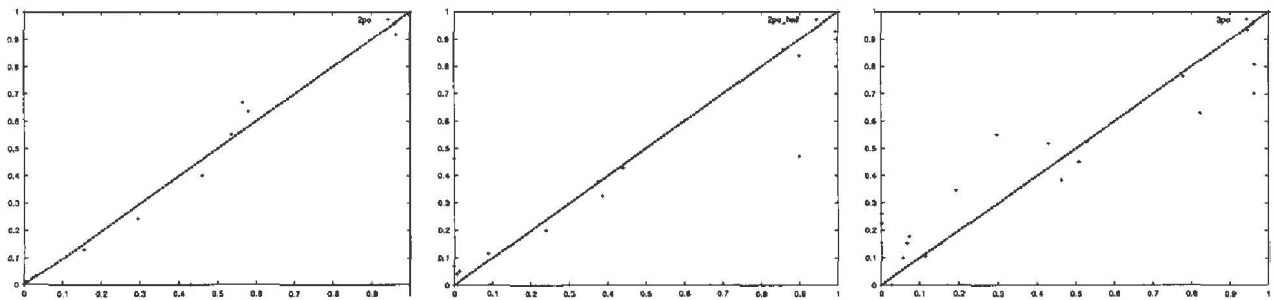

Figure 6: Comparing estimates of the posterior probabilities of child-parent features from two different MCMC runs. BNs restricted to those consistent with a specific ordering. For $k = 2$ (2po) runtimes=1472,1492; $k = 2$, half data (2po_half) runtimes = 1479,1447; and $k = 3$ (3po) runtimes=1960,1944.

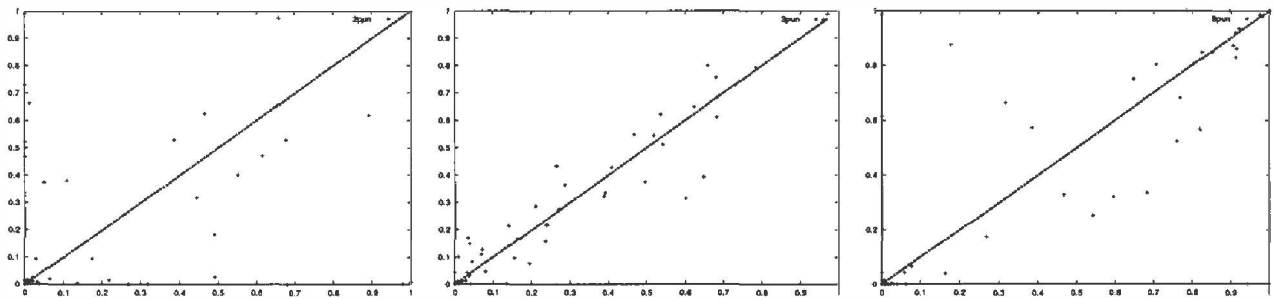

Figure 7: Comparing estimates of the posterior probabilities of child-parent features from two different MCMC runs. No variable ordering imposed. For $k = 2$ (2pun) runtimes=2273,2258; $k = 3$ (3pun) runtimes=2882,2891; and without any restriction on $k$ (8pun) runtimes=3340,3122

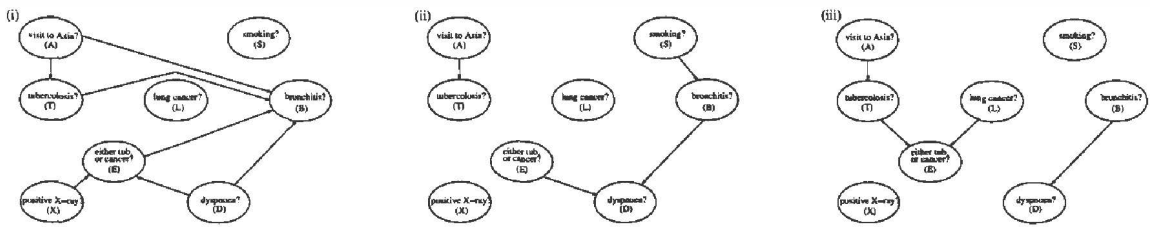

Figure 8: 'Recovered' BNs where only links with estimated posterior probability over 95% are created. For (i) the totally unconstrained case, (ii) when only a maximum of $k = 3$ parents were allowed and (iii) when only a maximum of $k = 3$ parents were allowed, but the parents of $E$ were forced to be $T$ and $L$. No variable ordering was imposed.